\documentclass[lettersize,journal]{IEEEtran}
\usepackage{amsmath,amsfonts}
\usepackage{algorithmic}
\usepackage{algorithm}
\usepackage{array}
\usepackage[caption=false,font=normalsize,labelfont=sf,textfont=sf]{subfig}
\usepackage{textcomp}
\usepackage{stfloats}
\usepackage{url}
\usepackage{verbatim}
\usepackage{graphicx}
\usepackage{cite}
\usepackage{times}
\usepackage{epsfig}
\usepackage{graphicx}
\usepackage{amssymb}
\usepackage{mathabx}
\usepackage{multirow}
\usepackage{fontawesome}
\usepackage{multicol}
\usepackage{multirow}

\usepackage{pifont}

\begin{document}

\title{Spatial and Frequency Domain Adaptive Fusion Network for Image Deblurring}

\author{Hu Gao, Depeng Dang$^\dag$
\thanks{Hu Gao and Depeng Dang are with the School of
Artificial Intelligence, Beijing Normal University,
Beijing 100000, China (e-mail: gao\_h@mail.bnu.edu.cn, ddepeng@bnu.edu.cn).

$\dag$ Corresponding Author
}

}

\maketitle

\begin{abstract}
Image deblurring aims to reconstruct a latent sharp image from its corresponding blurred one. Although existing methods have achieved good performance, most of them operate exclusively in either the spatial domain or the frequency domain, rarely exploring solutions that fuse both domains. In this paper, we propose a spatial-frequency domain adaptive fusion network (SFAFNet) to address this limitation. Specifically, we design a gated spatial-frequency domain feature fusion block (GSFFBlock), which consists of three key components: a spatial domain information module, a frequency domain information dynamic generation module (FDGM), and a gated fusion module (GFM). The spatial domain information module employs the NAFBlock to integrate local information. Meanwhile, in the FDGM, we design a learnable low-pass filter that dynamically decomposes features into separate frequency subbands, capturing the image-wide receptive field and enabling the adaptive exploration of global contextual information.  Additionally, to facilitate information flow and the learning of complementary representations. In the GFM, we present a gating mechanism (GATE)  to re-weight spatial and frequency domain features, which are then fused through the cross-attention mechanism (CAM).
Experimental results demonstrate that our SFAFNet performs favorably compared to state-of-the-art approaches on commonly used benchmarks. The code and the pre-trained models will be released at \url{https://github.com/Tombs98/SFAFNet}.
\end{abstract}

\section{Introduction}
Due to target movement, camera shake, and defocusing, images often become blurred. Image deblurring aims to recover high-quality images from their corrupted counterparts. Given the ill-posed nature of this inverse problem, many conventional approaches~\cite{2011Image,karaali2017edge} address this problem  based on various assumptions or hand-crafted features to constrain the solution space to natural images. However, designing such priors is challenging and lacks generalizability, making them impractical for real-world scenarios.

Significant progress in image deblurring has been achieved through the development of deep neural networks, which directly learn the mapping from the degraded observation to the clear image. Convolutional neural network (CNN)-based methods have become the leading choice for image deblurring by designing various architectures, including encoder-decoder architectures~\cite{MR-VNet,CascadedGaze}, multi-stage networks~\cite{Zamir2021MPRNet, Chen_2021_CVPR}, dual networks~\cite{2022Learning}, generative models~\cite{DBGAN,deganv2}, and more. While the convolution operation effectively models local connectivity, its limited receptive field and independence from input content hinder the model's ability to capture long-range dependencies. Unlike convolution operations that focus on local connectivity, Transformers excel at capturing non-local information. However, their attention mechanism introduces quadratic time complexity, leading to significant computational overhead. To mitigate this, some methods~\cite{Zamir2021Restormer,potlapalli2023promptir} adopt channel-wise self-attention instead of spatial dimensions, yet this sacrifices spatial information exploitation. Other approaches~\cite{Wang_2022_CVPR,u2former} employ non-overlapping window-based self-attention for single image deblurring, but struggle to fully harness information within each patch.

\begin{figure}
    \centering
    \includegraphics[width=1\linewidth]{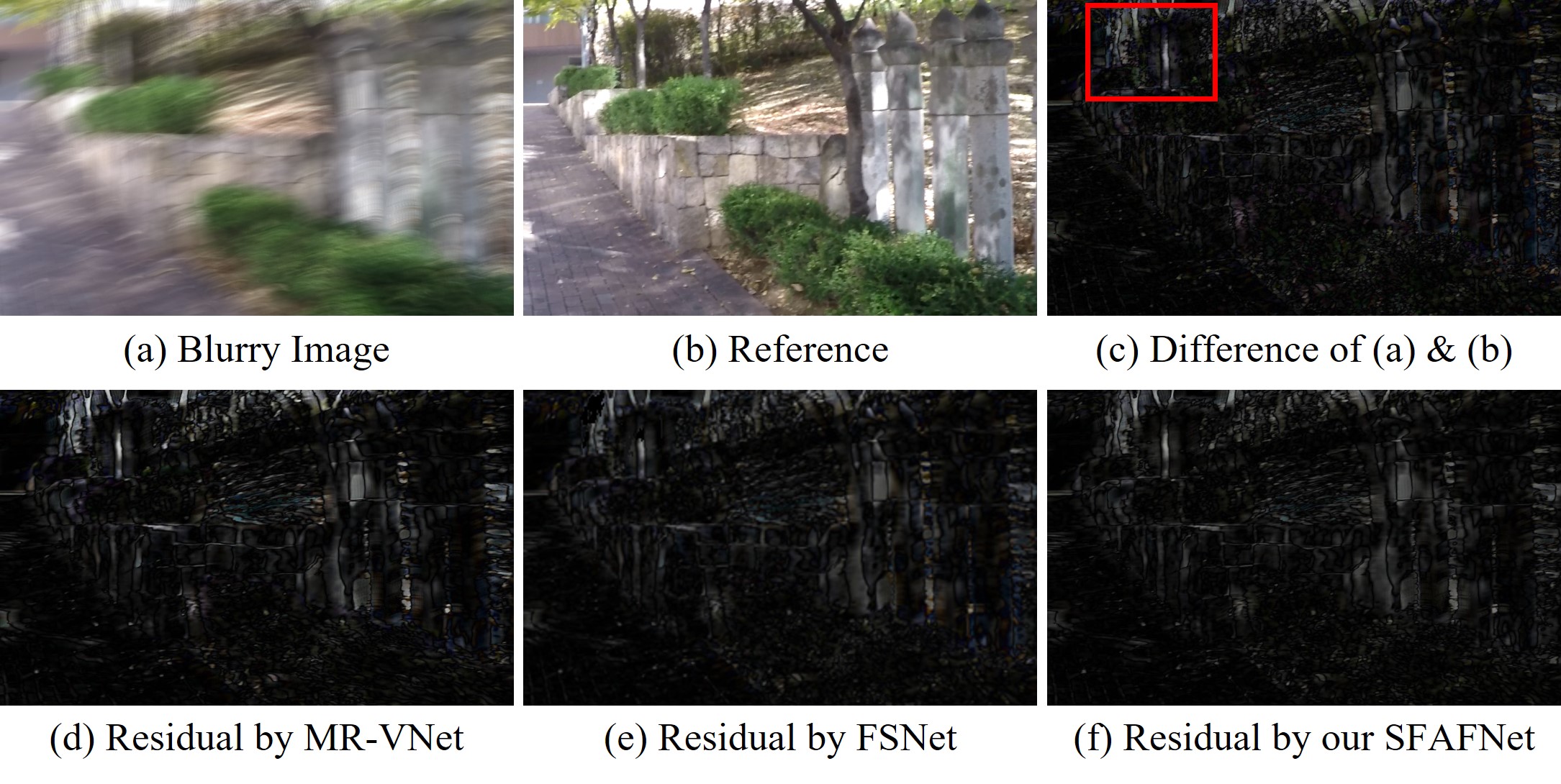}
    \caption{Visual comparison with MR-VNet~\cite{MR-VNet} and FSNet~\cite{FSNet}. The MR-VNet based on spatial domain often overlooks  details. Although the FSNet based on frequency captures details well, it struggles with spatially-variant properties. Our SFAFNet adaptively fuses spatial and frequency features, effectively learning detailed information and spatially-variant structures.}
    \label{fig:exam}
\end{figure}

The aforementioned methods primarily focus on spatial domain restoration, which often overlooks the frequency disparities between sharp and degraded image pairs. To address this challenge, a few approaches~\cite{kong2023efficient,fLi2023ICLR,AdaRevD} leverage transformation tools to explore the frequency domain. Nonetheless, these methods typically require Fourier inverse or wavelet transforms, leading to additional computational overhead and limited flexibility in selecting fusion frequency information. To effectively choose the most informative frequency components for reconstruction,~\cite{FSNet, SFNet, 10196308, 10747495} emphasize or attenuate resulting frequency components. However, we find that these frequency domain-based methods often neglect to effectively capture the spatial variation property of the overall structure.

As shown in Figure~\ref{fig:exam}, we visualize the results of several deblurring methods by displaying the residual image, which is the difference between the restored image and the degraded image. Figure~\ref{fig:exam}(c) shows the residual image of the ground truth image and the blurred image. It is evident that blurring affects not only the spatial structure but also results in a loss of detail (e.g., the trees enclosed in the red boxes). The spatial domain deblurring method MR-VNet~\cite{MR-VNet} overlooks details and introduces errors in the spatial structure (Figure~\ref{fig:exam}(d)). Although FSNet~\cite{FSNet} explores the most informative components for recovery from a frequency perspective to obtain detail-rich images, this approach does not adequately model the spatial variation of the overall structure (Figure~\ref{fig:exam}(e)). This limitation may lead to blurred false structures in the final deblurred image.

\begin{figure}
    \centering
    \includegraphics[width=1\linewidth]{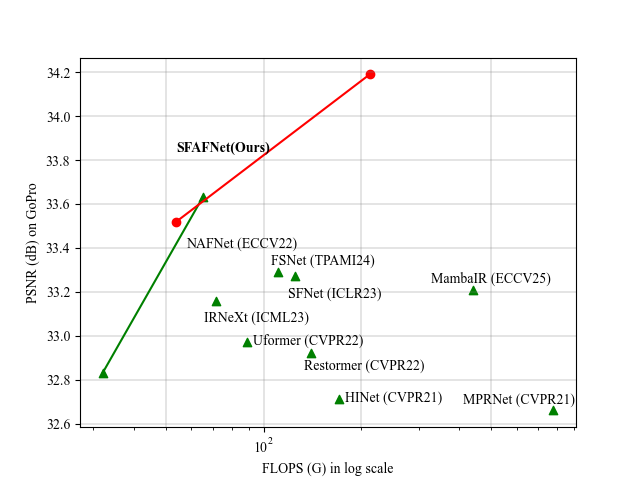}
    \caption{Computational cost vs. PSNR of models on the GoPro dataset~\cite{Gopro}. Our SFAFNet achieves the SOTA performance.}
    \label{fig:param}
\end{figure}

Based on the above analysis, we introduce a novel perspective for image deblurring. Specifically, we first attempt to address this problem in the spatial-frequency domain and propose the spatial-frequency domain adaptive fusion network, called SFAFNet. To implement SFAFNet, we design the fundamental building block named GSFFBlock, comprising three essential components: the spatial domain information module, the frequency domain information dynamic generation module (FDGM), and the gated fusion module (GFM). In the spatial domain information module, instead of designing novel modules, we use the NAFBlock \cite{chen2022simple} to integrate local information. Meanwhile, in the FDGM, we design a learnable low-pass filter that dynamically decomposes features into separate frequency subbands, capturing the image-wide receptive field and enabling the adaptive exploration of global contextual information.
Finally, to facilitate accurate information flow and the learning of complementary representations, the GFM first introduces a gating mechanism (GATE) to re-weight spatial and frequency domain features, ensuring that the most important information is preserved. Following this, we develop a  effective cross-attention mechanism (CAM) to integrate the features after the GATE process. We perform comprehensive experiments to validate the efficacy of the proposed networks, showcasing their remarkable performance superiority compared to state-of-the-art approaches across two prominent image deblurring tasks: image motion deblurring and image defocus deblurring. As illustrated in Figure~\ref{fig:param}, our SFAFNet model achieves SOTA performance  on the GoPro dataset~\cite{Gopro} compared to existing methods.

The main contributions are summarized as follows:
\begin{enumerate}
	\item We explore potential solutions for image deblurring in both the spatial and frequency domains and propose the spatial-frequency domain adaptive fusion network (SFAFNet). Extensive experiments demonstrate that SFAFNet achieves state-of-the-art performance.
 
    \item  We introduce a gated spatial-frequency domain feature fusion block (GSFFBlock), which consists of a spatial domain information module, a frequency domain information dynamic generation module (FDGM), and a gated fusion module (GFM). It facilitates the learning of complementary representations between local spatial information and global frequency information, thereby significantly enhancing the model’s learning capacity.
    
    \item We design the FDGM to dynamically decompose features into separate frequency subbands using a theoretically proven learnable low-pass filter.
    
    \item We present the GFM, which first re-weights spatial and frequency domain features via a gating mechanism (GATE), and then integrates these features through a  effective cross-attention mechanism (CAM).
\end{enumerate}

\section{Related Work}
\subsection{Hand-crafted prior-based methods.}
Due to the ill-posed nature of image deblurring, many conventional approaches~\cite{2011Image,karaali2017edge} address this challenge by incorporating hand-crafted priors to constrain the set of possible solutions. While these priors can aid in blur removal, the image degradation process is often uncertain. Consequently, these methods not only face difficulties in accurately modeling the degradation process but also often lack generalization ability.

\begin{figure*}
    \centering
    \includegraphics[width=0.9\linewidth]{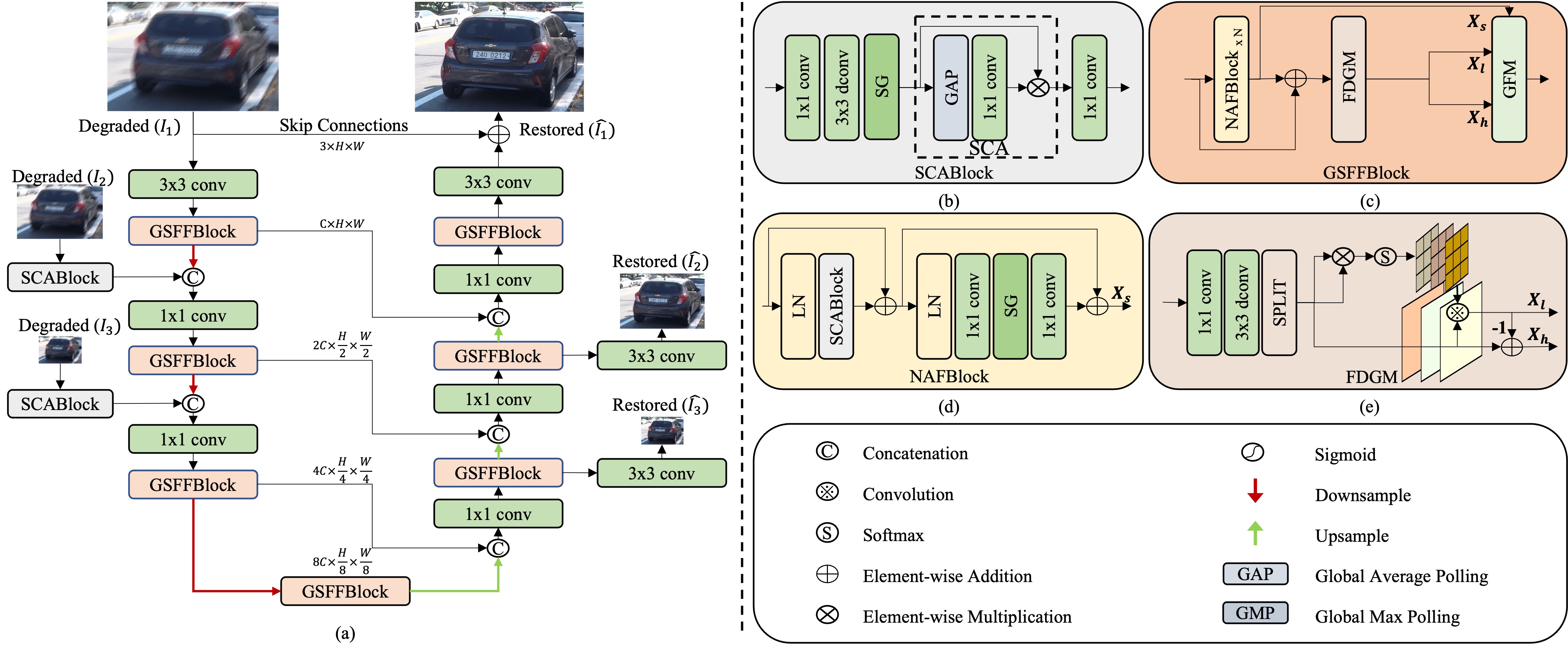}
    \caption{(a) Overall architecture of the proposed SFAFNet. (b) Simplified channel attention block (SCABlock) extracts shallow features.   (c) Gated spatial-frequency domain feature fusion block (GSFFBlock), which consists of N NAFBlocks~\cite{chen2022simple}, a frequency domain information dynamic generation module (FDGM), and a gated fusion module (GFM). (d) NAFBlock used to extract spatial domain features. (e) FDGM dynamically decompose features into separate frequency subbands.}
    \label{fig:network}
\end{figure*}
\subsection{Spatial-based methods.}
With the rapid advancement of deep learning, instead of manually designing image priors, many methods have turned to developing various deep CNN architectures. MPRNet~\cite{Zamir2021MPRNet} decomposes the image restoration process into manageable steps to maintain spatial details and contextual informations.
MIRNet-V2~\cite{Zamir2022MIRNetv2} introduces a multi-scale architecture that ensures spatially-precise representations are maintained , while also gathering complementary contextual information.
NAFNet~\cite{chen2022simple} analyzes baseline modules and demonstrates that non-linear activation functions may be dispensable.
CGNet~\cite{CascadedGaze} employs a global context extractor to effectively capture global context information.
MR-VNet~\cite{MR-VNet} utilizes the Volterra layers to optimally introduce non-linearities in the restoration process. 
Nonetheless,  the inherent limitations of convolutional operations restrict the models' ability to effectively remove long-range degradation artifacts.

To address these challenges, Transformers~\cite{2017Attention} have been applied to image deblurring. However, traditional Transformer architectures face significant computational overhead.
To improve efficiency, SwinIR~\cite{liang2021swinir} and U$^2$former~\cite{u2former} adopt window-based self-attention mechanisms in Transformer architectures. Additionally, Restormer~\cite{Zamir2021Restormer} and PromptIR~\cite{potlapalli2023promptir} compute self-attention across channels rather than spatial dimensions, resulting in linear complexity and enhancing computational efficiency.
While these methods achieve better performance than  hand-crafted approaches, they predominantly emphasize the spatial domain and often neglect the frequency differences between sharp and degraded image pairs.

\subsection{Frequency-based methods.}
Based on the spectral convolution theorem and the frequency disparities between sharp and degraded image pairs, it is feasible to process different frequency subbands individually in the frequency domain, effectively capturing global information.  Considering these advantages, several frequency-based methods have been proposed for image deblurring.
SDWNet~\cite{SDWNet} proposes a wavelet reconstruction module to recover more high-frequency details. FFTformer~\cite{kong2023efficient} leverages the convolution theorem to explore the properties of Transformers. DeepRFT~\cite{fxint2023freqsel} integrates Fourier transform to incorporate kernel-level information into image deblurring networks. AdaRevD~\cite{AdaRevD} introduces a FourierBlock to decode blur patterns. However, these methods typically require Fourier inverse or wavelet transforms, leading to additional computational overhead and limited flexibility.

To effectively choose the most informative frequency components for reconstruction, FocalNet~\cite{focalnetcui2023focal} and IRNeXt~\cite{IRNeXt} design conventional filters to generate different frequency signals. DDANet~\cite{DDAnet} devises a frequency attention module that performs controlled frequency transformation. AirFormer~\cite{10196308} proposes a supplementary prior module to selectively filter. MRLPFNet~\cite{MRLPFNet} exploits a learnable low-pass filter module to adaptively explore the global contexts.
SFNet~\cite{SFNet} and FSNet~\cite{FSNet} utilize multi-branch and content-aware modules to dynamically and locally decompose features into separate frequency subbands.

Nonetheless, these frequency domain-based methods often neglect to effectively capture the spatial variation property.  In this paper, we propose the spatial-frequency domain adaptive fusion network to  facilitate the learning of complementary representations between local spatial information and global frequency information.

\section{Method}
In this section, we first provide an overview of the entire SFAFNet pipeline. Then, we detailthe frequency domain information dynamic generation module (FDGM), and the gated fusion module (GFM) within the gated spatial-frequency domain feature fusion block (GSFFBlock). 

\subsection{Overall Pipeline} 
Our proposed SFAFNet, depicted in Figure~\ref{fig:network}, follows the widely accepted encoder-decoder architecture for effective hierarchical representation learning~\cite{gao2024learning}. Each encoder-decoder consists of a GSSFBlock, which further contains three key components. As shown in Figure~\ref{fig:network}(c), our GSFFBlock first captures spatial features $X_s$ using NAFBlock~\cite{chen2022simple}, then dynamically decomposes features into separate frequency subbands via FDGM. Finally, these spatial and frequency features are fused by GFM. The entire process can be defined as:
\begin{equation}
\begin{aligned}
	\label{equ:gssf}
   &X_s = NAFBlock_1(...(NAFBlock_N(X_{in})...)\\
   &X_l, X_h = FDGM(X_{in} \oplus X_s)\\
   &X_{out}  =  GFM(X_s, X_l, X_h)
\end{aligned}
\end{equation}

Given a degraded image $\mathbf{I} \in \mathbb{R}^{H \times W \times 3}$, SFAFNet first applies convolution to extract shallow features $\mathbf{F_{0}} \in \mathbb{R}^{H \times W \times C}$ (where $H$, $W$, and $C$ represent the height, width, and number of channels of the feature map, respectively). These shallow features pass through a three-scale encoder sub-network. It is important to note the use of multi-input and multi-output mechanisms to enhance training. Low-resolution degraded images are incorporated into the main path via SCABlock (refer to Figure~\ref{fig:network}(b)) and concatenation, followed by convolution for channel adjustment. The deepest features are then fed into a three-scale decoder. Throughout this process, the encoder features are concatenated with the decoder features to aid in reconstruction. Finally, we refine the features to generate a residual image $\mathbf{I_R} \in \mathbb{R}^{H \times W \times 3}$, which is added to the degraded image to produce the restored image: $\mathbf{\hat{I}} = \mathbf{I_R} + \mathbf{I}$. 

To facilitate the selection process across both spatial and frequency domains, we optimize the proposed network SFAFNet with the following loss function:
\begin{equation}
\begin{aligned}
\label{eq:loss1}
L &= \sum_{i=1}^{4}(L_{c}(\hat{I_i},\overline I_i)  + \delta L_{e}(\hat{I_i},\overline I_i) + \lambda L_{f}(\hat{I_i},\overline I_i))
\\
L_{c} &= \sqrt{||\hat{I_i} -\overline I_i||^2 + \epsilon^2}
\\
L_{e} &= \sqrt{||\triangle \hat{I_i} - \triangle \overline I_i||^2 + \epsilon^2}
\\
L_{f} &= ||\mathcal{F}(\hat{I}_i)-\mathcal{F}(\overline I_i)||_1
\end{aligned}
\end{equation}
where $i$ denotes the index of input/output images at different scales, $\overline I_i$ denotes the target images and $L_{c}$ is the  Charbonnier loss with constant $\epsilon$ empirically set to $0.001$ for all the experiments. $L_{e}$ is the edge loss, where $\triangle$ represents the  Laplacian operator. $L_{f}$  denotes the frequency domains loss, where $\mathcal{F}$ represents fast Fourier transform, and the parameters $\lambda$ and $\delta$ control the relative importance of loss terms, which are set to $0.1$ and $ 0.05$ as in~\cite{Zamir2021MPRNet,FSNet}, respectively.

\subsection{Spatial Domain Information Module}
This study aims to explore the adaptive fusion of spatial and frequency domain features to achieve accurate image deblurring, focusing on both detail information and spatial structure. Consequently, instead of designing a new spatial domain feature capture module, we utilize the existing NAFBlock~\cite{chen2022simple}. Figure~\ref{fig:network}(d) illustrates the process of obtaining spatial features $X_s$ from an input $X$ using Layer Normalization (LN), simplified channel attention block (SCABlock), and Simple Gate (SG). Express as follows:
\begin{equation}
\begin{aligned}
	\label{equ:0xnaf}
    X_1 &= SCABlock(LN(X)) + X \\
    X_s &= X_1 + f_{1 \times 1}^{c}(SG(f_{1 \times 1}^{c}(LN(X_1))))\\
\end{aligned}
\end{equation}
where $f_{1 \times 1}^{c}$ denotes the $1 \times 1$ convolution,  SG represents the simple gating mechanism, which starts by splitting a feature into two along the channel dimension. Then computes these two features using a linear gate. SCABlock is shown in Figure~\ref{fig:network}(b), we also use it to extract shallow features for low-resolution input images. The process of SCABlock can be defined as :
\begin{equation}
\begin{aligned}
	\label{equ:0naf}
	X_1^{'} &= f_{1 \times 1}^{c}(SCA(SG(f_{3 \times 3}^{dwc}(f_{1 \times 1}^{c}(X^{'}))),\\
    SCA &= X_0^{'} \cdot f_{1 \times 1}^{c}(GAP(X_0^{'}))
\end{aligned}
\end{equation}
where $f_{3 \times 3}^{dwc}$ denotes the $3 \times 3$ depth-wise convolution, GAP is the global average pool.

\subsection{Frequency Domain Information Dynamic Generation Module}
\label{ros}
Existing methods primarily focus on spatial domain restoration, often neglecting the frequency differences between sharp and degraded image pairs. While some methods use transformation tools like wavelet transform to obtain frequency domain information, these approaches lack flexibility. To address this issue, we analyze the low-pass filter. A low-pass filter performs a weighted average over an area the same size as the filter, thereby filtering out high-frequency information and retaining low-frequency components. To be specific, given an image $I_i$ and a low-pass filter $LF$ with the filter weight $W_{xy}$ (where $W_{xy} \ge 0, \sum_y W_{xy} = 1$, $x, y$ denote image pixels),  we can obtain a filtered image by:
\begin{equation}
	\label{equ:lf}
LF(I_i(x))  = \sum_y W_{xy} I_i(y)
\end{equation}

Leveraging this property of low-pass filters, we design a frequency domain information dynamic generation module (FDGM) to decompose features into separate frequency subbands dynamically and locally using a learnable low-pass filter. Specifically, given any spatial feature map $F$, we first generates three projections$F^1, F^2, F^3$, enriched with spatial context.
Then, we leverage two of these projections to produce the low-pass filter for each row of the input. The process can be defined as:

\begin{equation}
\begin{aligned}
	\label{equ:fl}
	F^1, F^2, F^3 &= SPLIT(f_{3 \times 3}^{dwc}(f_{1 \times 1}^{c}(F)))\\
    F^L &= Softmax(F^1 \otimes F^2)
\end{aligned}
\end{equation}
where SPLIT is the operation of divide feature map by channel in thirds. $F^L \in \mathbb{R}^{r \times k \times k }$, $r$ is the number of row, and $k \times k$ denotes the kernel size of low-pass filters. The Softmax normalized operation is applied to each row to ensure $F^L_{xy} \ge 0$ and $\sum_y F^L_{xy} = 1$. Thus, each $r$ of $F^L$ can be regarded as a low-pass filter.  
Compared to the fixed-mode low-pass filter and the learnable MDSF proposed by FSNet~\cite{FSNet}, our learnable filter is based on the correlation metric of data components. This makes it spatially variant and allows it to model the main structures more adaptively.

To attain the high-pass filter, we subtract the resulting low-pass filter from the identity kernel with the central value as one and everywhere else as zero. Next, for each row feature $F_i \in \mathbb{R}^{H \times W \times \frac{C}{r} }$
, where i is the row index, its low- and high-frequency components can
be obtained by using the corresponding filter $F^L$ and $F^H$, which is expressed as:
\begin{equation}
\begin{aligned}
	\label{equ:flh}
	X_l &=  F^L \ocoasterisk F^3_{i,x,y,c}\\
 X_h &= F^H  \ocoasterisk F^3_{i,x,y,c}
\end{aligned}
\end{equation}
where $c$ is the index of  channel and $ \ocoasterisk$ represents the convolution.

\begin{figure}
    \centering
    \includegraphics[width=1\linewidth]{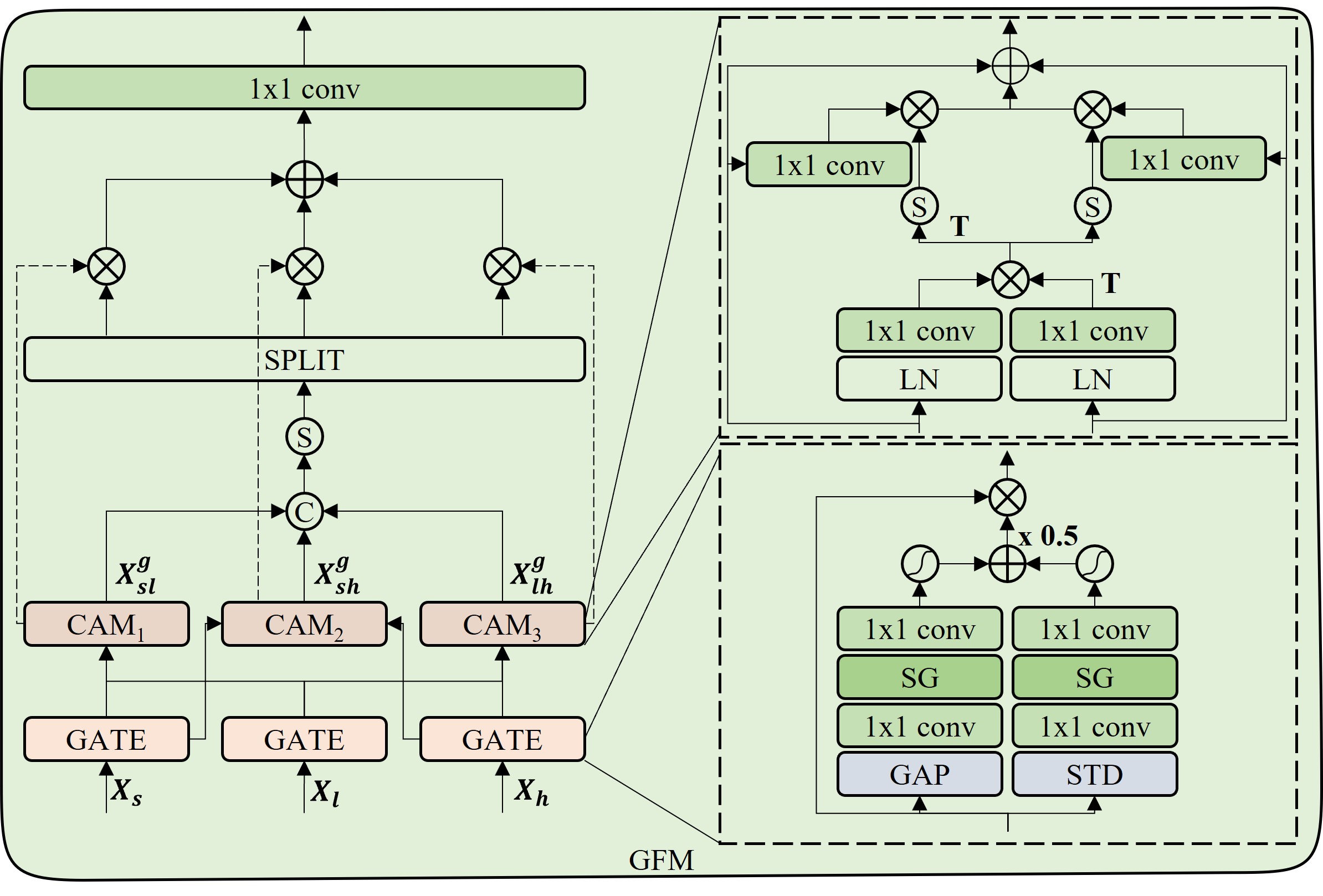}
    \caption{Gated fusion module (GFM) contains gating mechanism (GATE) and cross-attention mechanism (CAM).}
    \label{fig:gfm}
\end{figure}

In FDGM, we use convolution to apply the filter to the input feature map within a $k \times k$ region, where the value of the central pixel is the sum of all re-weighted pixel values in the region. $k \times k$, some pixels contributing to the edge values are outside this region. Next, based on the above operators, we prove the validity of the low-pass filter. The proof is similar to SFNet~\cite{SFNet}. 

\textbf{Theorem 1.} Given $W, F \in \mathbb{R}^{k^2 \times k^2}, W_i = Sof tmax(F_i), i = 0, 1, ..., k^2$ - $ 1$. Then $W$ is a low-pass filter. For all $m \in \mathbb{R}^{k^2}$, $lim_{p \to \infty} \frac{|| HF[W^pm] ||_2}{|| W^pm ||_2} = 0$. Where $HF[\cdot]$ is the high-frequency operator, which can be described as:
\begin{equation}
\begin{aligned}
	\label{equ:kkpf}
	HF[I_i] &= F_{k^2}^{-1}diag(0,1,...1)F_{k^2}I_i
          \\
          & = F_{k^2}^{-1}(E-diag(1,0,...0))F_{k^2}I_i
          \\
          &= (E-\frac{1}{k^2}M)I_i
\end{aligned}
\end{equation}
where $F$ denotes the discrete Fourier transform, $E$ is the unit matrix, diag(·) denotes the diagonal matrix, and $M$ represents the  matrix with all values as 1.

\textbf{Proof.} Note that after Softmax operation, $W_{xy} \ge 0, \sum_y W_{xy} = 1$. 
Thus $W$ has an eigenvalue  $\lambda_1 = 1$, and the corresponding eigenvector is $v = [1,1,...,1]$. Let scalars $\lambda_2,..., \lambda_s$ denote the other eigenvalues of $W$, and they all are less than $\lambda_1$, according to Perron-Frobenius Theorem~\cite{meyer2023matrix}. 

The Jordan canonical form $W^pm$ can be writer as:
\begin{equation}
\begin{aligned}
	\label{equ:pf}
	&W^pm = TJ^pT^{-1}m \\
 &= \begin{pmatrix}a_1 & ... &a_{k^2}\end{pmatrix}
 \begin{pmatrix} \lambda_1^p & &  
 \\ & J_2(\lambda_2)^p & &
 \\& & \ddots &
 \\
 & & &J_s(\lambda_s)^p
 \end{pmatrix}
 \begin{pmatrix}b_1\\ \vdots \\b_{k^2}\end{pmatrix}
 m
\end{aligned}
\end{equation}
where $T$ is the  transition matrix, $a, b \in \mathbb{R}^{k^2}, a_1 = v, J(\lambda) \in \mathbb{R}^{z_i \times z_i}, i \in [2,s]$ is the  Jordan block, which can be computer by:
\begin{equation}
	\label{equ:pddf}
	J_i(\lambda_i) = \begin{pmatrix}
 \lambda_i^p & (\lambda_i^p)^{'} & \cdots & \frac{(\lambda_i^p)^{z_i-1}}{(z_i-1)!}
 \\ & \ddots & \ddots & \vdots
 \\ & & \ddots& (\lambda_i^p)^{'}
 \\ & & & \lambda_i^p
\end{pmatrix}_{z_i}
\end{equation}

For arbitrary $z_i \ge q \ge 1$ and $|\lambda_i| \textless 1$, due to the the larger growing rate of exponential function  than that of power function. Therefore, except the main diagonal values 
$lim_{p \to \infty}	\frac{p(p-1)...(p-1+1)}{q!} \lambda_i^{p-q} = 0 $.
As a consequence,  $lim_{p \to \infty}W_pm = Tdiag(\lambda_1^t,0,...,0)T^{-1}m = vb_1m$. And the the theorem 1 can be computed as:
\begin{equation}
\begin{aligned}
	\label{equ:phhjf}
	lim_{p \to \infty} \frac{|| HF[W^pm] ||_2}{|| W^pm ||_2} &= lim_{p \to \infty} \frac{|| (E-\frac{1}{k^2}M)W^pm ||_2}{|| W^pm ||_2}
\\ &=  lim_{p \to \infty} \frac{|| (Evb_1m - \frac{1}{k^2} Mvb_1m ||_2}{|| W^pm ||_2} 
\\ & =0
\end{aligned}
\end{equation}

\subsection{Gated Fusion Module}

Our primary objective is to explore potential solutions for image deblurring in both the spatial and frequency domains. With this goal in mind, we capture spatial domain features $X_s$ and frequency domain features $X_l, X_h$. The key challenge now is how to adaptively fuse these features to enhance the model's learning capacity. To achieve this goal, we design the gated fusion module (GFM) to facilitate accurate information flow and the learning of complementary representations via the gating mechanism (GATE) and the cross-attention mechanism (CAM), as illustrated in Figure~\ref{fig:gfm}. The GATE  re-weight spatial and frequency domain features, ensuring that the most important information is preserved. The CAM  integrate the features after the GATE process.  Next, we provide details of both process. 

\textbf{GATE.}  The GATE consists of two branches. In the left branch, we calculate the mean of each feature channel is firstly obtained by performing global average pooling, while in the lower branch, we compute the standard deviation of input features by global standard deviation pooling (STD). Following this, the respective statistic vectors in two branches are
fed into two fully connected layers, where SG and Sigmoid activation functions are followed after each layer, respectively. Subsequently, the output from the two branches are averaged to obtain the re-weight coefficients. Finally, these re-weight coefficients are multiplied by the input features to obtain the re-weighted features $X_s^g, X_l^g, X_h^g$. 

\textbf{CAM.} These re-weighted features $X_s^g, X_l^g, X_h^g$ are pairwise fused using CAM. For simplicity, we use the fusion of $X_s^g$ and $X_h^g$ as an example. First, layer normalization is applied, followed by a $1 \times 1$ convolution layer to derive the feature. Next, a similarity matrix $X_{sh}^{gs}$ is obtained by computing the dot product. Two Softmax functions are then applied on each row of the similarity matrix to obtain two attention score matrices. These attention matrices are used to weight the original features, resulting in the matrices $X_{sh}^{gs}, X_{sh}^{gh}$ that contain mutual information. Finally, these matrices are summed to obtain the cross-attention features $X_{sh}^{g}$, which fuse the spatial and frequency domain features. The process of CAM is formulated as follows:

\begin{equation}
\begin{aligned}
	\label{equ:cam}
	X_{sh}^{gs} &= f_{1 \times 1}^{c}(LN(X_s^g)) \otimes (f_{1 \times 1}^{c}(LN(X_h^g)))^T
 \\
    X_{sh}^{gs} &= f_{1 \times 1}^{c}(X_s^g) \otimes Softmax((X_{sh}^{ga})^T)
    \\
    X_{sh}^{gh} &= f_{1 \times 1}^{c}(X_h^g) \otimes Softmax((X_{sh}^{ga}))
    \\
    X_{sh}^{g} &= X_s^g \oplus X_{sh}^{gr} \oplus X_h^g \oplus X_{sh}^{gl}
\end{aligned}
\end{equation}
where $T$ denotes transpose. These features  $X_{sh}^{g}$ can be treated as containing spatial domain and high-frequency features. In the same way, we can also obtain other cross-attention features $X_{sl}^{g}, X_{lh}^{g}$. Then, these cross-attention features $X_{sl}^{g}, X_{sh}^{g}, X_{lh}^{g}$ are further weighted  to facilitate information flow and the learning of complementary representations as follows:

\begin{equation}
\begin{aligned}
	\label{equ:cam2}
	&X_{sl}^{gw}, X_{sh}^{gw}, X_{lh}^{gw} =  SPLIT(Softmax([X_{sl}^{g}, X_{sh}^{g}, X_{lh}^{g}]))\\
 & X_{sf} =  f_{1 \times 1}^{c}(X_{sl}^{g} \otimes X_{sl}^{gw} \oplus X_{sh}^{g} \otimes X_{sh}^{gw} \oplus   X_{lh}^{g} \otimes X_{lh}^{gw} )
\end{aligned}
\end{equation}
where $[\cdot]$ represents the channel-wise concatenation, and $ X_{sf}$ is the feature that adaptively fuses both spatial and frequency domain features to enhance the model's learning capacity.

\section{Experiments}
In this section, we describe the experimental settings and subsequently present both qualitative and quantitative comparisons between SFAFNet and other state-of-the-art methods. Following this, we conduct ablation studies to validate the effectiveness of our approach. The best and second-best results in tables are highlighted in \textbf{bold} and \underline{underlined} formats, respectively.

\subsection{Experimental Settings}


\subsubsection{\textbf{Training details}}
We train separate models for different tasks, and unless otherwise specified, the following parameters are utilized.  The models are trained using the Adam optimizer~\cite{2014Adam} with parameters $\beta_1=0.9$ and $\beta_2=0.999$. The initial learning rate is set to $2 \times 10^{-4}$ and gradually reduced to $1 \times 10^{-6}$ using the cosine annealing strategy~\cite{2016SGDR}. The batch size is chosen as $32$, and patches of size $256 \times 256$ are extracted from training images. Data augmentation involves horizontal and vertical flips. $N$ (see Figure~\ref{fig:network} (c)) is set to 15, and the number of rows is 8 (see Section~\ref{ros}). Due to the enormous complexity of image motion deblurring,  we set the number of channels to 32 for SFAFNet and 64 for SFAFNet-B. 

\begin{figure*}[hb] 
	\centering
	\includegraphics[width=1\linewidth]{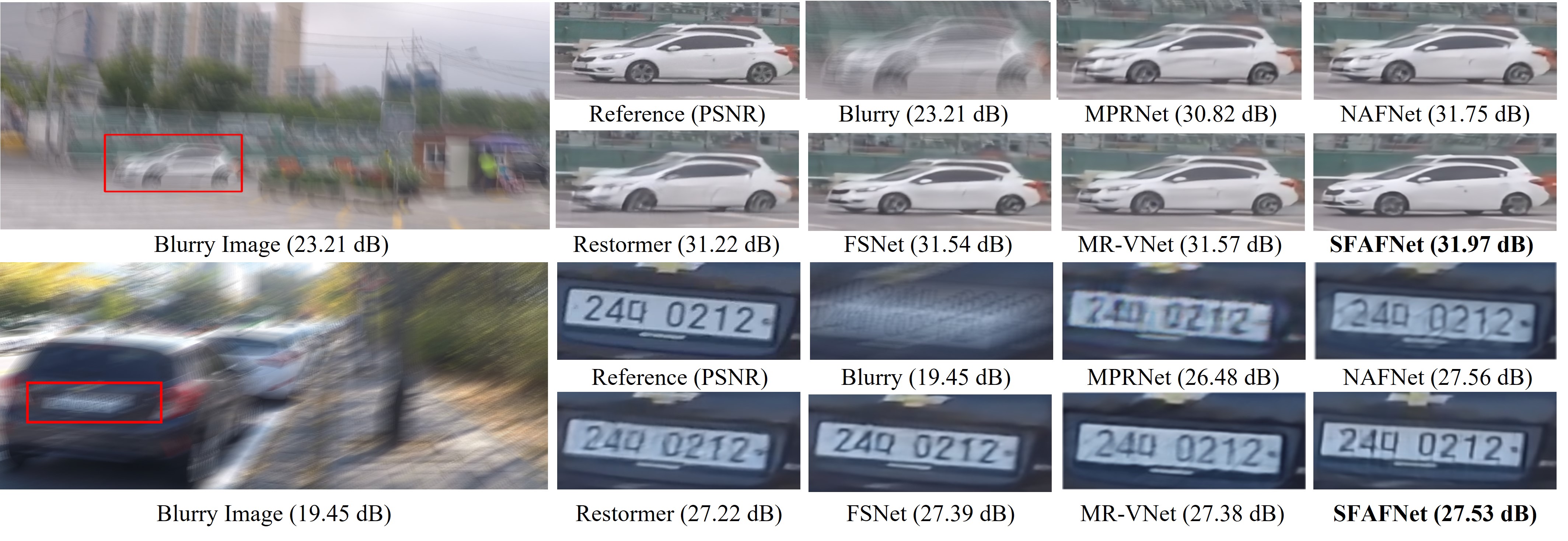}
	\caption{\textbf{Image motion deblurring} comparisons on the GoPro dataset~\cite{Gopro}. Our SFAFNet recovers perceptually faithful images. }
	\label{fig:blurm}
\end{figure*}

\subsubsection{Datasets}
In this part, we introduce the datasets used for two prominent image deblurring tasks: image motion deblurring and image defocus deblurring. 

\textbf{Image Motion Deblurring.} 
We use the GoPro dataset~\cite{Gopro} to verify the effectiveness of our method, following recent approaches~\cite{FSNet}. This dataset includes 2,103 image pairs for training and 1,111 pairs for evaluation. To assess the generalizability of our approach, we directly apply the GoPro-trained model to the test images of the HIDE~\cite{HIDE} dataset, which contains 2,025 images specifically collected for human-aware motion deblurring. Both the GoPro and HIDE datasets are synthetically generated.
To evaluate the performance of our method on real-world images, we further test on the RealBlur~\cite{realblurrim_2020_ECCV} dataset, which includes 3,758 image pairs for training and 980 image pairs for testing. This dataset comprises two subsets: RealBlur-J and RealBlur-R.

\textbf{Single-Image Defocus Deblurring.}
To evaluate the effectiveness of our method, we adopt the DPDD dataset~\cite{DPDNet}, following the approach of recent methods~\cite{FSNet}. This dataset includes images from 500 indoor and outdoor scenes captured with a DSLR camera. Each scene comprises three defocused input images (right view, left view, and center view) and a corresponding all-in-focus ground-truth image. The DPDD dataset is divided into training, validation, and testing sets, consisting of 350, 74, and 76 scenes, respectively. SFAFNet is trained using the center view images as input, with loss values computed between the outputs and the corresponding ground-truth images.

\begin{table}
\centering
\caption{Quantitative evaluations of the proposed approach against state-of-the-art motion deblurring methods. Our SFAFNet and SFAFNet-B are trained only on the GoPro dataset. \label{tb:deblurgh}}
\resizebox{\linewidth}{!}{
\begin{tabular}{ccccc}
    \hline
    \multicolumn{1}{c}{} & \multicolumn{2}{c}{GoPro}  & \multicolumn{2}{c}{HIDE} 
    \\
   Methods & PSNR $\uparrow$ & SSIM $\uparrow$ & PSNR $\uparrow$ & SSIM $\uparrow$   
    \\
    \hline\hline
    MPRNet~\cite{Zamir2021MPRNet} & 32.66 & 0.959 & 30.96 & 0.939 
    \\
    HINet~\cite{Chen_2021_CVPR}&32.71&0.959&30.32&0.932
    \\
    Restormer~\cite{Zamir2021Restormer} & 32.92 & 0.961 & 31.22 & 0.942 
    \\
     Uformer~\cite{Wang_2022_CVPR} &32.97 & 0.967 &30.83 &\textbf{0.952} 
     \\
       NAFNet-32~\cite{chen2022simple}&32.83&0.960&-&-
    \\
    NAFNet-64~\cite{chen2022simple}&33.62&0.967&-&-
    \\
    IRNeXt~\cite{IRNeXt} &33.16 &0.962 &- & - 
    \\
    SFNet~\cite{SFNet} &33.27 &0.963 &31.10 & 0.941 
    \\
    DeepRFT+~\cite{fxint2023freqsel} &  33.23 &0.963 &\underline{31.66} &0.946
    \\
    DDANet~\cite{DDAnet} &33.07 &0.962 &30.64 &0.937
    \\
    MRLPFNet~\cite{MRLPFNet} &34.01 &0.968 &31.63 &0.947
    \\
      DeblurDiNAT-L~\cite{DeblurDiNAT}&33.63 &0.967 &31.47 &0.944
     \\
     MambaIR~\cite{guo2024mambair}&33.21 &0.962 &31.01 &0.939
     \\
     MR-VNet~\cite{MR-VNet} & \underline{34.04} & \underline{0.969} & 31.54 & 0.943
     \\
    FSNet~\cite{FSNet} &33.29&0.963 &31.05 & 0.941 
    \\
    \hline
    \textbf{SFAFNet(Ours)}& 33.52	&0.964	&31.62	&0.946

    \\
    \textbf{SFAFNet-B(Ours)} & \textbf{34.25} & \textbf{0.971} & \textbf{31.92} & \underline{0.949}
    \\
    \hline
\end{tabular}}
\end{table}

\begin{table}
\centering
\caption{ Quantitative evaluations of the proposed approach against
state-of-the-art methods on the RealBlur  dataset~\cite{realblurrim_2020_ECCV}. All the comparison results are generated using the publicly available codes and
the models trained on the same training datasets.
}
\label{tb:0deblurringreal}
\resizebox{\linewidth}{!}{
\begin{tabular}{ccccc}
    \hline
    \multicolumn{1}{c}{} & \multicolumn{2}{c}{RealBlur-R}  & \multicolumn{2}{c}{RealBlur-J} 
    \\
   Methods & PSNR $\uparrow$ & SSIM $\uparrow$ & PSNR $\uparrow$ & SSIM $\uparrow$   
    \\
    \hline\hline
    DeblurGAN-v2~\cite{deganv2} & 36.44 & 0.935& 29.69& 0.870
\\
    MPRNet~\cite{Zamir2021MPRNet} & 39.31 & 0.972 & 31.76 & 0.922
   \\
   DeepRFT+~\cite{fxint2023freqsel}&39.84 &0.972 &32.19 &0.931
\\
Stripformer~\cite{Tsai2022Stripformer} & 39.84 & 0.975 & 32.48 & 0.929
\\
FFTformer~\cite{kong2023efficient}&40.11& 0.973 &32.62 &0.932
\\
 MambaIR~\cite{guo2024mambair}& 39.92 & 0.972 & 32.44 & 0.928
 \\
 MR-VNet~\cite{MR-VNet} & \underline{40.23} & \underline{0.977} & \underline{32.71} & \underline{0.941}
 \\
 \hline
    \textbf{SFAFNet(Ours)}& \textbf{40.83} & \textbf{0.980} & \textbf{32.95} & \textbf{0.947}
    \\
    \hline
\end{tabular}}
\end{table}

\begin{table*}[htb]
    \centering
        \caption{Quantitative comparisons with other single-image defocus deblurring methods on the DPDD testset~\cite{DPDNet} (containing 37 indoor and 39 outdoor scenes).}
    \label{tab:deblurd}
    \resizebox{\linewidth}{!}{
    \begin{tabular}{c|cccc|cccc|cccc}
        \hline
    \multicolumn{1}{c|}{} & \multicolumn{4}{c|}{Indoor Scenes}  & \multicolumn{4}{c|}{Outdoor Scenes} & \multicolumn{4}{c}{Combined}
    \\
   Methods & PSNR $\uparrow$ & SSIM $\uparrow$ &MAE $\downarrow$  &LPIPS $\downarrow$ & PSNR $\uparrow$ & SSIM $\uparrow$  &MAE $\downarrow$  &LPIPS $\downarrow$  &  PSNR $\uparrow$ &  SSIM $\uparrow$ &MAE $\downarrow$  &LPIPS $\downarrow$ 
    \\
    \hline
    \hline
KPAC~\cite{KPAC}& 27.97 &0.852 &0.026 &0.182 &22.62 &0.701 &0.053 &0.269 &25.22 &0.774 &0.040 &0.227
\\
IFAN~\cite{IFAN}& 28.11 &0.861 &0.026 &0.179 &22.76 &0.720 &0.052 &0.254 &25.37 &0.789 &0.039 &0.217
\\
Restormer~\cite{Zamir2021Restormer}& 28.87 &\underline{0.882} &0.025 &\textbf{0.145} &23.24 &0.743 &0.050 &\textbf{0.209} &25.98 &0.811 &0.038 &\textbf{0.178}
\\
IRNeXt~\cite{IRNeXt} &\underline{29.22} &0.879 &0.024 &0.167 &\underline{23.53} &\underline{0.752} &\underline{0.049} &0.244 &\underline{26.30} &\underline{0.814} &\underline{0.037} &0.206
\\
SFNet~\cite{SFNet} &29.16 &0.878 &\underline{0.023} &0.168 &23.45 &0.747 &\underline{0.049} &0.244 &26.23 &0.811 &\underline{0.037} &0.207
\\
FSNet~\cite{FSNet} &29.14 &0.878 &0.024 &0.166 &23.45 &0.747 &0.050 &0.246 &26.22 &0.811 &\underline{0.037} &0.207
\\
MambaIR~\cite{guo2024mambair}& 28.89 &0.879 &0.026 &0.171 &23.36 &0.738 &0.051 &0.243 &26.11 &0.809 &0.039 &0.202
\\
MR-VNet~\cite{MR-VNet}& 29.19 &0.878 &0.025 &0.158 &23.46 &0.742 &0.051 &0.245 &26.29 &0.812 &0.037 &0.201
\\
\hline
\textbf{SFAFNet(Ours)}&\textbf{29.74}	&\textbf{0.908}	&\textbf{0.021}	&\underline{0.152}	&\textbf{24.01}	&\textbf{0.753}	&\textbf{0.045}	&\underline{0.230}	&\textbf{26.79}	&\textbf{0.815}	&\textbf{0.034}	&\underline{0.192}

    \\
    \hline
    \end{tabular}}
\end{table*}

\subsection{Experimental Results}
\subsubsection{ \textbf{Image Motion Deblurring}}
We present the performance of evaluated image deblurring approaches on the synthetic GoPro~\cite{Gopro} and HIDE~\cite{HIDE} datasets in Tables~\ref{tb:deblurgh}. 
Overall, our SFAFNet  generates higher-quality images with higher PSNR and SSIM values than the competing approaches.
Specifically, compared to our baseline network NAFNet~\cite{chen2022simple}, we improve 0.69 dB and 0.63 dB at 32  and 64  number of channels, respectively.  Compared with previous best spatial-based method, MR-VNet~\cite{MR-VNet}, our SFAFNet-B improve 0.21 dB on the GoPro~\cite{Gopro} dataset and 0.38 dB on the HIDE~\cite{HIDE} dataset.  In addition, our SFAFNet-B provides a significant gain of 0.96 dB over the frequency-based method FSNet~\cite{FSNet} on the GoPro~\cite{Gopro} dataset and 0.87 dB on the HIDE~\cite{HIDE} dataset.
Noted that, even though our network is trained solely on the GoPro~\cite{Gopro} dataset, it still achieves state-of-the-art results (31.92 dB in PSNR) on the HIDE~\cite{HIDE} dataset. This demonstrates its impressive generalization capability. Figure.~\ref{fig:blurm} shows some of the images deblurred by the evaluation method, our model produces visually more pleasing results.

\begin{figure*}[htb] 
	\centering
	\includegraphics[width=1\linewidth]{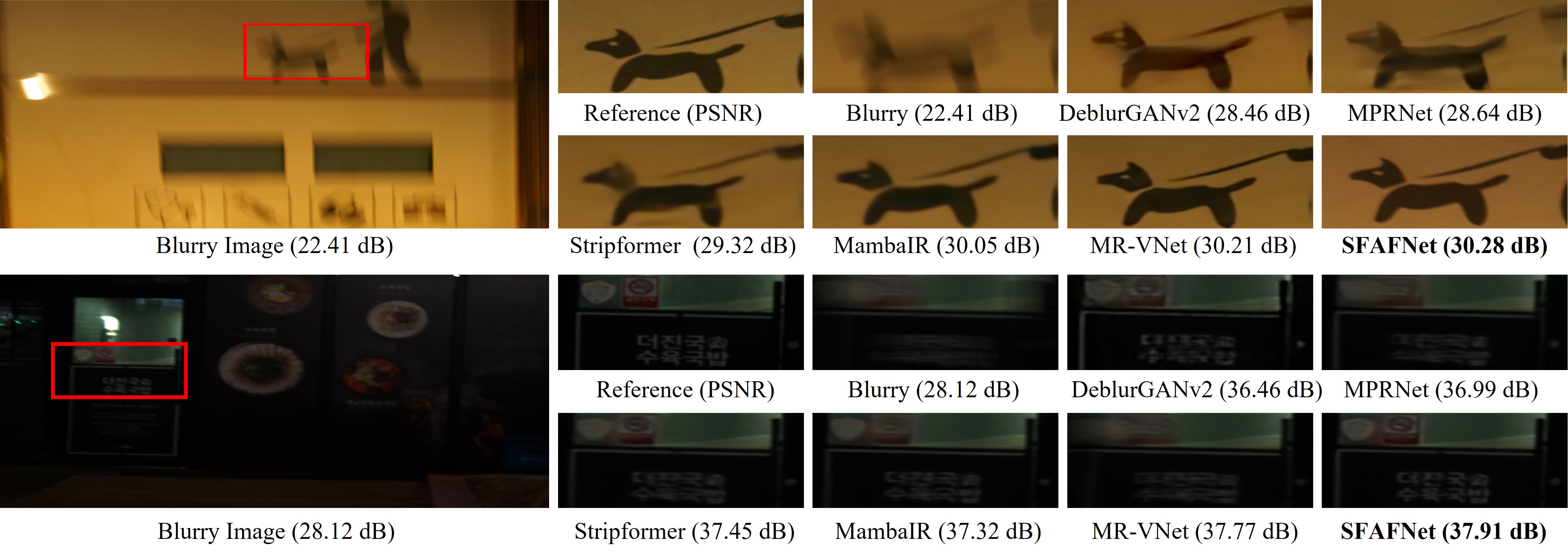}
	\caption{\textbf{Image motion deblurring} comparisons on the RealBlur dataset~\cite{realblurrim_2020_ECCV}. Our SFAFNet  recovers image with clearer details.}
	\label{fig:blurmreal}
\end{figure*}

In addition to the synthetic datasets, we further evaluate the effectiveness of our SFAFNet on real-world images from the RealBlur dataset~\cite{realblurrim_2020_ECCV}.
As shown in Table~\ref{tb:0deblurringreal},   the proposed method generates the deblurred results with higher PSNR and SSIM values. Specifically, compared with the previous best method MR-VNet~\cite{MR-VNet}, our gains are 0.60 dB and 0.24 dB on RealBlur-R and RealBlur-J, respectively. 
Figure~\ref{fig:blurmreal} illustrates that the proposed SFAFNet is able to explore both highand low-frequency information, generating a better image with clearer details and structures than the competed methods.

\begin{figure*}[htb] 
	\centering
	\includegraphics[width=1\textwidth]{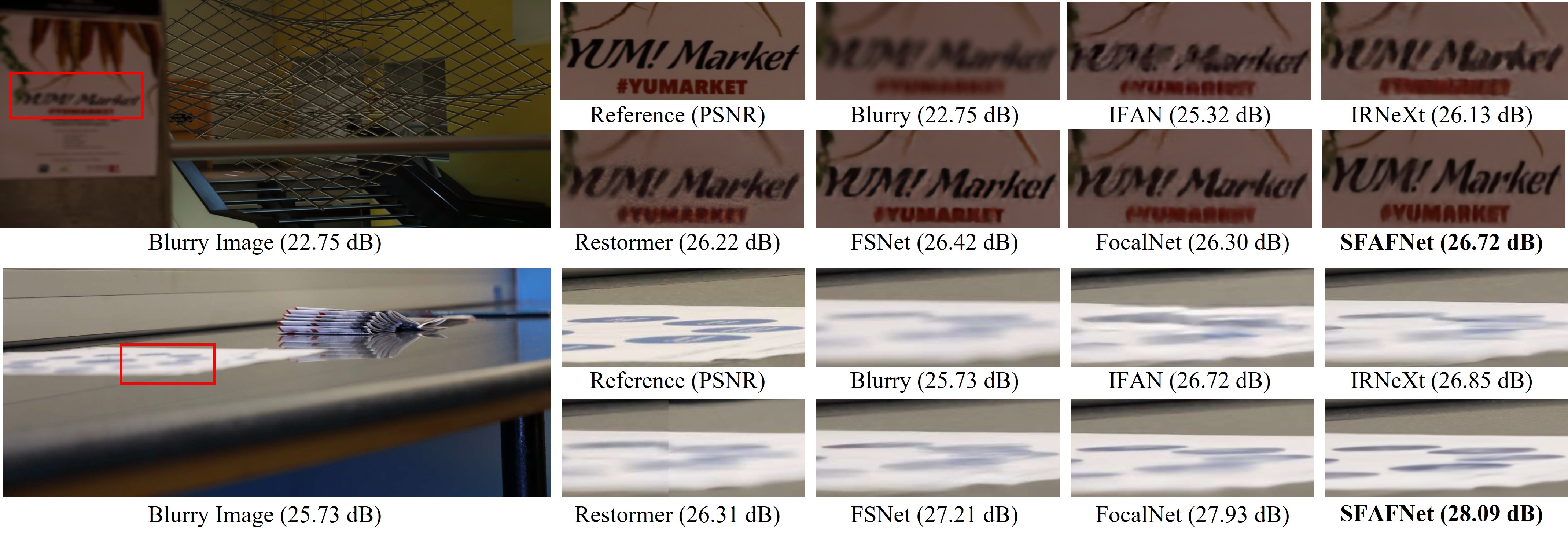}
	\caption{\textbf{Single image defocus deblurring} comparisons on the DDPD dataset~\cite{DPDNet}. Compared to the state-of-the-art methods, our SFAFNet effectively removes blur while preserving the fine image details. }
	\label{fig:blurd}
\end{figure*}

\subsubsection{\textbf{Single-Image Defocus Deblurring}}
We conduct single-image defocus deblurring experiments on the DPDD~\cite{DPDNet} dataset. Table~\ref{tab:deblurd} presents  image fidelity scores of state-of-the-art defocus deblurring methods. SFAFNet outperforms other state-of-the-art methods across all scene categories. Notably, in the combined scenes category, SFAFNet exhibits a 0.49 dB improvement over the leading frequency-based method IRNeXt~\cite{IRNeXt}. In addition, our method improves 0.50 dB over the spatial-based MR-VNet~\cite{MR-VNet}.
The visual results in Figure~\ref{fig:blurd} demonstrate that our method recovers more details and aligns more closely with the ground truth compared to other algorithms.

\subsection{Ablation Studies}
In this section, we first demonstrate the effectiveness of the proposed modules and then investigate the effects of different designs for each module. 


\begin{table}
    \centering
       \caption{Ablation study on individual components of the
proposed SFAFNet.}
    \label{tab:abl1}
    \resizebox{\linewidth}{!}{
    \begin{tabular}{cccccc}
    \hline
         \multirow{2}{*}{Net}&\multirow{2}{*}{FDGM}& \multicolumn{2}{c}{GFM}  & \multirow{2}{*}{PSNR} & \multirow{2}{*}{$\triangle$ PSNR}
         \\
         & & GATE&CAM & & 
         \\
         \hline
         (a)& &  &   & 32.77 & -
         \\
         (b)& \ding{52}&  &   & 32.92 & +0.15
         \\
         (c)&\ding{52} &  \ding{52}&  &  33.08 & +0.31
         \\
         (d)& \ding{52}&  &  \ding{52}&   33.29 & +0.52 
         \\
         (e)& \ding{52}&  \ding{52}&  \ding{52}&  33.52 & +0.75
         \\
         \hline
    \end{tabular}}
\end{table}

\begin{figure*}[htb] 
	\centering
	 \includegraphics[width=1\linewidth]{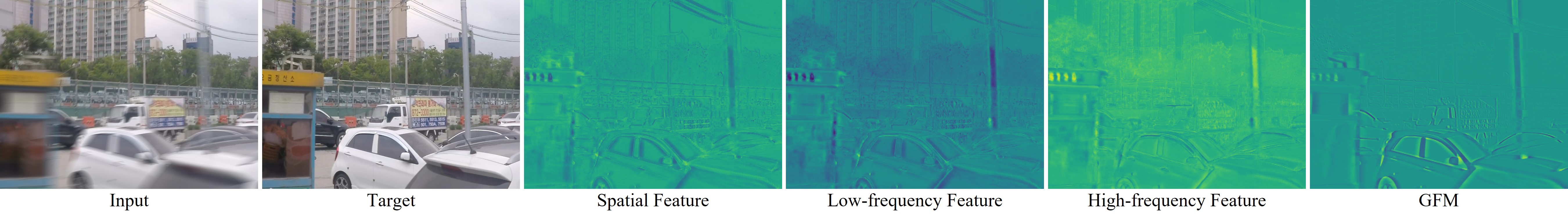}
	\caption{The internal features of GFM. Our GFM produces more fine details, such as the advertising on the car. Zoom in for the best view. }
	\label{fig:gfmv}
\end{figure*}
\subsubsection{Effects of Individual Components} 
As shown in Table~\ref{tab:abl1}(a), the baseline achieves a PSNR of 32.77 dB. Adding FDGM to the baseline and concatenating the resulting frequency domain features with spatial domain features improves the result by 0.15 dB (Table~\ref{tab:abl1}(b)). Using the GATE mechanism of GFM to reweight before concatenation further improves performance from 32.92 dB to 33.08 dB (Table~\ref{tab:abl1}(c)).  When the CAM in GFM is employed to fuse frequency and spatial domain features, the performance increases by 0.37 dB (Table~\ref{tab:abl1}(b) to Table~\ref{tab:abl1}(d)).
Regardless of the combination of these three modules, each yields a corresponding performance enhancement. When all modules are utilized together (Table~\ref{tab:abl1}(e)), our model achieves a 0.75 dB boost over the original baseline (Table~\ref{tab:abl1}(a)).

\begin{table}
    \centering
       \caption{Results of alternatives to FDGM. }
    \label{tab:ablfdgm}
    
    \begin{tabular}{ccc}
    \hline
         Method& PSNR  & FLOPs(G)
         \\
         \hline
         (a) Gaussian &  32.95 & 54.56
         \\
         (b) Wavelet & 32.93  & 54.29
         \\
         \hline
         (c) LPF~\cite{MRLPFNet} & 33.26  &  55.79
         \\
         (d) MDSF~\cite{FSNet} & 33.31 & 55.12
         \\
        (e) FDGM & 33.52  & 54.32
         \\
         \hline
    \end{tabular}
\end{table}

\subsubsection{Alternatives to FDGM} 
The FDGM dynamically decomposes features into separate frequency subbands, capturing the image-wide receptive field and enabling adaptive exploration of global contextual information. To demonstrate the effectiveness of FDGM, we first compare our method with fixed frequency separation methods: Gaussian(Table~\ref{tab:ablfdgm} (a)) and Wavelet (Table~\ref{tab:ablfdgm} (b)). Both achieve similar results, which are lower than our FDGM. Since our filter kernel is generated through learning, we further compare FDGM with two learnable-filter approaches: LPF~\cite{MRLPFNet} (Table~\ref{tab:ablfdgm} (c)) and MDSF~\cite{FSNet} (Table~\ref{tab:ablfdgm} (d)). The result shows that our method has clear advantages over these approaches, demonstrating FDGM's effectiveness. We further visualize the feature maps after the FDGM
in Figure~\ref{fig:gfmv}. Using the learnable low-pass filters, the FDGM produces low- and high-frequency components.

\subsubsection{The row number of FDGM}
As shown in Table~\ref{tab:aldfsg}, the performance improves as the number of rows increases, reaching an optimal value at 8 rows. Beyond this point, further increasing the number of rows leads to performance degradation due to overfitting.

\begin{table}
    \caption{The impact of row number on the overall performance.}
    \label{tab:aldfsg}
    \centering
    \begin{tabular}{ccccc}
    \hline
         Row&   4 & 6 & 8 & 12
         \\
         \hline
         PSNR&  33.44  &33.47  &33.52 & 33.49
         \\ 
         \hline
    \end{tabular}
\end{table}
 
\begin{table}
    \centering
       \caption{Results of alternatives to GFM. }
    \label{tab:ablgfm}
    
    \begin{tabular}{cccc}
    \hline
         Method& PSNR & $\triangle$ PSNR & FLOPs(G)
         \\
         \hline
         (a) - &    33.29 & - & 54.32
         \\
         (b) GFFB~\cite{tao2024hierarchical}& 33.39& +0.15 & 56.45
         \\
         (c) CC~\cite{latticenet9847064} &  33.41 &+0.17& 54.32
         \\
         (d) GATE & 33.52 &+0.23    &54.32
         \\
         \hline
         (e) Concatenate & 33.08 &- &   51.93
         \\
         (f) SKFF~\cite{Zamir2022MIRNetv2} & 33.35&+0.27  & 57.82
         \\
         (g) FM~\cite{IRNeXt} & 33.29 &+0.21  & 50.98
        \\
        (h) CAM & 33.52 &+0.44&  54.32
         
         \\
         \hline
    \end{tabular}
\end{table}

\begin{figure}[htb] 
	\centering
	 \includegraphics[width=1\linewidth]{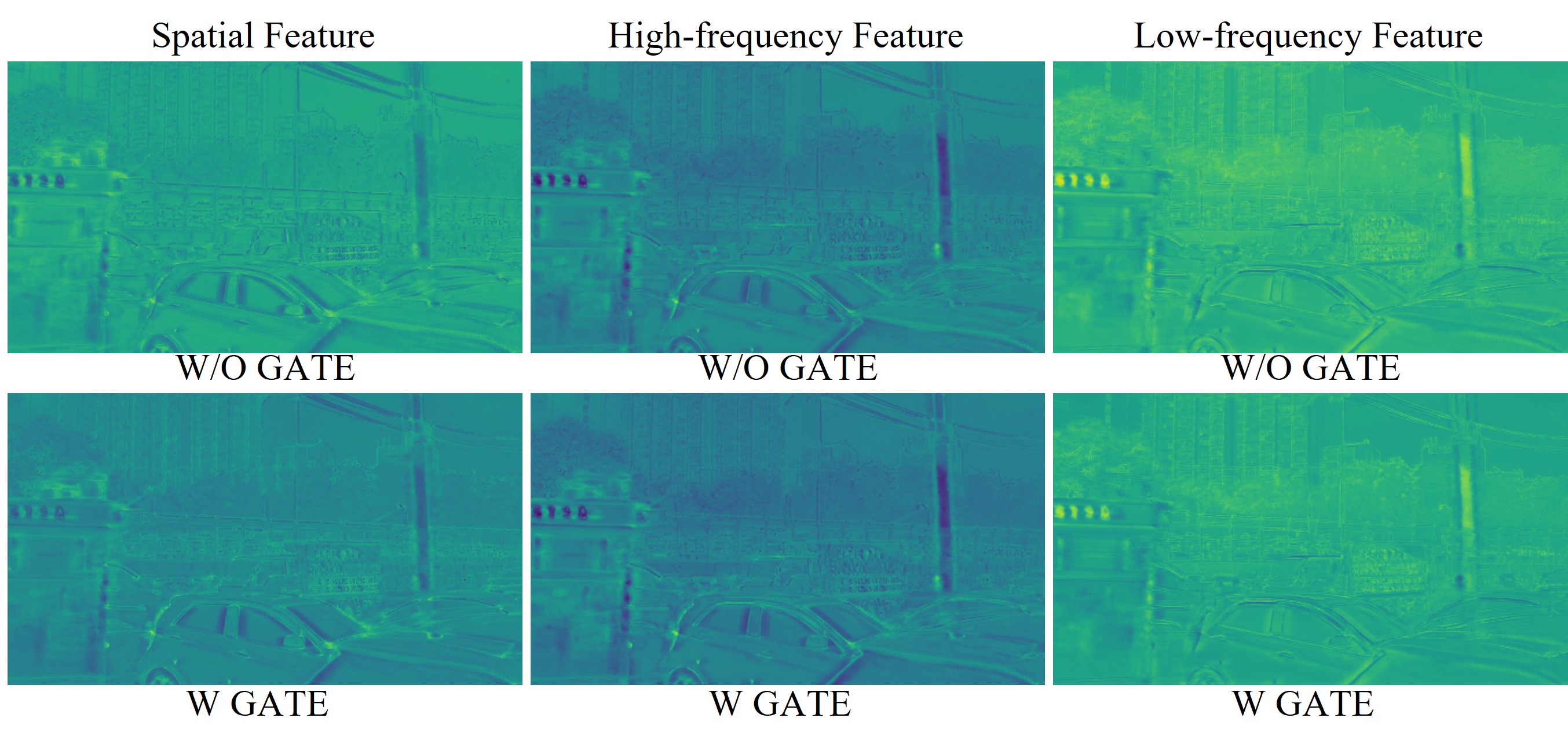}
	\caption{Visualization of intermediate feature maps of models
with and without GATE. }
	\label{fig:gate}
\end{figure}

\subsubsection{Design Choices for GFM} 
GFM improves performance as shown in Table~\ref{tab:abl1}. To examine the advantage of our design, we compare our GFM with several alternatives in Table~\ref{tab:ablgfm}. First, we verify the effectiveness of our GATE mechanism (Table~\ref{tab:ablgfm} (a)-(d)), with experimental results showing that our GATE achieves optimal results. Then, we replace CAM with an existing method for feature fusion(Table~\ref{tab:ablgfm} (e)-(h)), and the experimental results demonstrate that performance decreases when CAM is replaced. 

To delve into this mechanism, we  visualize the feature maps in Figure~\ref{fig:gfmv}. As expected, spatial features focus on capturing local texture details, while frequency features capture global information but lack spatial variation of the overall structure. Our GFM combines both spectral and spatial information, providing a comprehensive representation of the input image. These results indicate that the proposed dual-domain information integration mechanism effectively fuses complementary information from multiple domains, resulting in enhanced model performance. Furthermore, we plot resulting feature maps  of GFM for models with and without GATE in Figure~\ref{fig:gate}. As can be seen, equipped with GATE, the model recovers more accurate signals.

\begin{table}
\centering
\caption{The impact of loss function.}
\label{tb:loss}
\begin{tabular}{cccc}
    \hline
    $L_{char}$ & $L_{freq}$ & $L_{edge}$ & PSNR
    \\
    \hline
    &  & & 33.32
    \\
    \ding{52} &&  & 33.38
      \\
    \ding{52} & \ding{52} &  & 33.45
      \\
    \ding{52} & \ding{52} & \ding{52} & 33.52
    \\
    \hline
\end{tabular}
\end{table}

\subsubsection{Loss function}
To investigate the influence of the loss function, we conduct experiments with different combinations of loss functions. The results, summarized in  Table~\ref{tb:loss}, show that employing the proposed loss function yields optimal results.

\subsubsection{Model Complexity}
We assess the model complexity of our approach and state-of-the-art methods in terms of running time and FLOPs.  Table~\ref{tab:computational} shows that our SFAFNet model achieves SOTA performance while reducing computational costs, underscoring the efficiency and resource effectiveness of our method. Specifically,  we achieve a 0.23 dB improvement over the previous best approach, FSNet~\cite{FSNet}, with up to 51.2\% cost reduction. Moreover, as shown in Figure~\ref{fig:param}, through the scaling up of the model size, our SFAFNet achieves even better performance, highlighting the scalability of SFAFNet.

\begin{table}
    \centering
    \caption{The evaluation of model computational complexity on the GoPro dataset~\cite{Gopro}. The FLOPs are evaluated on image patches with the size of 256×256 pixels. The running time is evaluated on images with the size of 1280 × 720 pixels. All the results are obtained on a
machine with an NVIDIA GeForce RTX 1060 GPU.}
    \label{tab:computational}

    \begin{tabular}{cccc}
    \hline
         Method& Time(s) & FLOPs(G)  & PSNR 
         \\
         \hline\hline
         MPRNet~\cite{Zamir2021MPRNet} & 0.981 & 777.01 & 32.66 
         \\
         Restormer~\cite{Zamir2021Restormer} & 0.980 & 140.99 & 32.92 
         \\
         Stripformer~\cite{Tsai2022Stripformer} &0.659 &170.21 & 33.08 
         \\
         DeepRFT+~\cite{fxint2023freqsel} & 0.552& 187.04&  33.23 
         \\
         FSNet~\cite{FSNet} &\underline{0.365} & \underline{111.14}& \underline{33.29}
         \\
         MambaIR~\cite{guo2024mambair} & 0.743 & 439.36 & 33.21 
         \\
         \hline
         SFAFNet(Ours) &\textbf{0.358} &\textbf{54.32} & \textbf{33.52}
         \\
         \hline
    \end{tabular}
\end{table}

\section{Conclusion}
In this paper, we propose the spatial-frequency domain adaptive fusion network (SFAFNet) to address image deblurring in both the spatial and frequency domains. Specifically, we introduce a  gated spatial-frequency domain feature fusion block (GSFFBlock), which consists of a spatial domain information module, a frequency domain information dynamic generation module (FDGM), and a gated fusion module (GFM).  The FDGM dynamically decomposes features into separate frequency subbands using a theoretically proven learnable low-pass filter. The GFM re-weights spatial and frequency domain features via a gating mechanism (GATE) and then integrates these features through an effective cross-attention mechanism (CAM). Extensive experiments demonstrate that SFAFNet achieves state-of-the-art performance.

\bibliographystyle{IEEEtran}
\bibliography{aaai25}

\vfill

\end{document}